\documentclass[11pt]{article}
\usepackage{acl2016}
\usepackage{url}
\usepackage{times}
\usepackage{latexsym}
\usepackage{xcolor}
\aclfinalcopy % Uncomment this line for the final submission
\usepackage[utf8]{inputenc}
\usepackage[T1]{fontenc}

\usepackage{enumitem}
\usepackage{amsmath}
\usepackage{multirow}

\usepackage{siunitx}
\renewcommand\footnotemark{}
\title{Edinburgh Neural Machine Translation Systems for WMT 16}

\author{
Rico Sennrich\and Barry Haddow \and Alexandra Birch\\
School of Informatics, University of Edinburgh\\
{\tt \{rico.sennrich,a.birch\}@ed.ac.uk}, {\tt bhaddow@inf.ed.ac.uk}
}

\date{}

\begin{document}
\maketitle
\begin{abstract}

We participated in the WMT 2016 shared news translation task by building neural translation systems for four language pairs, each trained in both directions:
English$\leftrightarrow$Czech, English$\leftrightarrow$German, English$\leftrightarrow$Romanian and English$\leftrightarrow$Russian.
Our systems are based on an attentional encoder-decoder,
using BPE subword segmentation for open-vocabulary translation with a fixed vocabulary.
We experimented with using automatic back-translations of the monolingual News corpus as additional training data,
pervasive dropout, and target-bidirectional models.
All reported methods give substantial improvements, and we see improvements of 4.3--11.2 {\sc Bleu} over our baseline systems.
In the human evaluation, our systems were the (tied) best constrained system for 7 out of 8 translation directions in which we participated.\footnote{We have released the implementation that we used for the experiments as an open source toolkit: \url{https://github.com/rsennrich/nematus}}\footnote{We have released scripts, sample configs, synthetic training data and trained models: \url{https://github.com/rsennrich/wmt16-scripts}}

\end{abstract}

\section{Introduction}

We participated in the WMT 2016 shared news translation task by building neural translation systems for four language pairs:
English$\leftrightarrow$Czech, English$\leftrightarrow$German, English$\leftrightarrow$Romanian and English$\leftrightarrow$Russian.
Our systems are based on an attentional encoder-decoder \cite{DBLP:journals/corr/BahdanauCB14},
using BPE subword segmentation for open-vocabulary translation with a fixed vocabulary \cite{DBLP:journals/corr/SennrichHB15}.
We experimented with using automatic back-translations of the monolingual News corpus as additional training data \cite{2015arXiv151106709S},
pervasive dropout \cite{2015arXiv151205287G}, and target-bidirectional models.

\section{Baseline System}

Our systems are attentional encoder-decoder networks \cite{DBLP:journals/corr/BahdanauCB14}.
We base our implementation on the dl4mt-tutorial\footnote{\url{https://github.com/nyu-dl/dl4mt-tutorial}}, which we enhanced with new features such as ensemble decoding and pervasive dropout.

We use minibatches of size 80, a maximum sentence length of 50, word embeddings of size 500, and hidden layers of size 1024.
We clip the gradient norm to 1.0 \cite{DBLP:conf/icml/PascanuMB13}.
We train the models with Adadelta \cite{DBLP:journals/corr/abs-1212-5701}, reshuffling the training corpus between epochs.
We validate the model every \num{10000} minibatches via {\sc Bleu} on a validation set (newstest2013, newstest2014, or half of newsdev2016 for EN$\leftrightarrow$RO).
We perform early stopping for single models, and use the 4 last saved models (with models saved every \num{30000} minibatches) for the ensemble results.
Note that ensemble scores are the result of a single training run.
Due to resource limitations, we did not train ensemble components independently, which could result in more diverse models and better ensembles.

Decoding is performed with beam search with a beam size of 12.
For some language pairs, we used the AmuNMT C++ decoder\footnote{\url{https://github.com/emjotde/amunmt}} as a more efficient alternative to the theano implementation of the dl4mt tutorial.

\subsection{Byte-pair encoding (BPE)}

To enable open-vocabulary translation, we segment words via byte-pair encoding (BPE)\footnote{\url{https://github.com/rsennrich/subword-nmt}} \cite{DBLP:journals/corr/SennrichHB15}.
BPE, originally devised as a compression algorithm \cite{Gage:1994:NAD:177910.177914}, is adapted to word segmentation as follows:

First, each word in the training vocabulary is represented as a sequence of characters, plus an end-of-word symbol.
All characters are added to the symbol vocabulary.
Then, the most frequent symbol pair is identified, and all its occurrences are merged, producing a new symbol that is added to the vocabulary.
The previous step is repeated until a set number of merge operations have been learned.

BPE starts from a character-level segmentation, but as we increase the number of merge operations, it becomes more and more different from a pure character-level model in that frequent character sequences, and even full words, are encoded as a single symbol.
This allows for a trade-off between the size of the model vocabulary and the length of training sequences.
The ordered list of merge operations, learned on the training set, can be applied to any text to segment words into subword units that are in-vocabulary in respect to the training set (except for unseen characters).

To increase consistency in the segmentation of the source and target text, we combine the source and target side of the training set for learning BPE.
For each language pair, we learn \num{89500} merge operations.

\section{Experimental Features}

\subsection{Synthetic Training Data}

WMT provides task participants with large amounts of monolingual data, both in-domain and out-of-domain.
We exploit this monolingual data for training as described in \cite{2015arXiv151106709S}.
Specifically, we sample a subset of the available target-side monolingual corpora, translate it automatically into the source side of the respective language pair,
and then use this synthetic parallel data for training.
For example, for EN$\to$RO, the back-translation is performed with a RO$\to$EN system, and vice-versa.

\newcite{2015arXiv151106709S} motivate the use of monolingual data with domain adaptation, reducing overfitting, and better modelling of fluency.
We sample monolingual data from the News Crawl corpora\footnote{Due to recency effects, we expect last year's corpus to be most relevant, and sampled from News Crawl 2015 for EN-RO, EN-RU and EN-CS; for EN-DE, we re-used data from \cite{2015arXiv151106709S}, which was randomly sampled from News Crawl 2007--2014.}, which is in-domain with respect to the test set.

\begin{table}
\centering
\begin{tabular}{l|rrrr}
type & DE & CS & RO & RU\\
\hline
parallel & 4.2 & 52.0 & 0.6 & 2.1 \\
synthetic ($*\to$EN) & 4.2 & 10.0 & 2.0 & 2.0\\
synthetic (EN$\to*$) & 3.6 & 8.2 & 2.3 & 2.0\\
\end{tabular}
\caption{Amount of parallel and synthetic training data (number of sentences, in millions) for EN-* language pairs. 
%For synthetic data, original language of monolingual data is listed in parentheses.
For synthetic data, we separate the data according to whether the original monolingual language is English or not. 
}
\label{data}
\end{table}

The amount of monolingual data back-translated for each translation direction ranges from 2 million to 10 million sentences.
Statistics about the amount of parallel and synthetic training data are shown in Table \ref{data}.
With dl4mt, we observed a translation speed of about \num{200000} sentences per day (on a single Titan X GPU).

\subsection{Pervasive Dropout}

For English$\leftrightarrow$Romanian, we observed poor performance because of overfitting.
To mitigate this, we apply dropout to all layers in the network, including recurrent ones. 

Previous work dropped out different units at each time step.
When applied to recurrent connections, this has the downside that it impedes the information flow over long distances,
and \newcite{DBLP:conf/icfhr/PhamBKL14} propose to only apply dropout to non-recurrent connections.

Instead, we follow the approach suggested by \newcite{2015arXiv151205287G}, and use the same dropout mask at each time step.
Our implementation differs from the recommendations by \newcite{2015arXiv151205287G} in one respect: we also drop words at random, but we do so on a token level, not on a type level.
In other words, if a word occurs multiple times in a sentence, we may drop out any number of its occurrences, and not just none or all.

In our English$\leftrightarrow$Romanian experiments, we drop out full words (both on the source and target side) with a probability of 0.1.
For all other layers, the dropout probability is set to 0.2.

\subsection{Target-bidirectional Translation}

We found that during decoding, the model would occasionally assign a high probability to words based on the target context alone, ignoring the source sentence.
We speculate that this is an instance of the label bias problem \cite{lafferty2001conditional}.

To mitigate this problem, we experiment with training separate models that produce the target text from right-to-left (r2l),
and re-scoring the n-best lists that are produced by the main (left-to-right) models with these r2l models.
Since the right-to-left model will see a complementary target context at each time step, we expect that the averaged probabilities will be more robust.
In parallel to our experiments, this idea was published by \newcite{liu2016}.

We increase the size of the n-best-list to 50 for the reranking experiments.

A possible criticism of the l-r/r-l reranking approach is that the gains actually come from adding diversity to the ensemble, since we
are now using two independent runs. However experiments in \cite{liu2016} show that a l-r/r-l reranking systems is
stronger than an ensemble created from two independent l-r runs.

\section{Results}

\subsection{English$\leftrightarrow$German}

\begin{table}
\centering
\begin{tabular}{l|cc|cc}
system & \multicolumn{2}{c|}{EN$\to$DE} & \multicolumn{2}{c}{DE$\to$EN}\\
& dev & test & dev & test\\
\hline
baseline & 22.4 & 26.8 &%/mnt/hoenir0/rsennrich/wmt16_neural/en-de/exp10/model.npz.npz.best_bleu
26.4 & 28.5\\%/mnt/hoenir0/rsennrich/wmt16_neural/de-en/exp3/model.npz.npz.best_bleu
+synthetic & 25.8 & 31.6&%/mnt/hoenir0/rsennrich/wmt16_neural/en-de/exp1/model.npz.npz.best_bleu
29.9 & 36.2\\ %/mnt/hoenir0/rsennrich/wmt16_neural/de-en/exp1/model.npz.npz.best_bleu
+ensemble & 27.5 & 33.1 &%/mnt/hoenir0/rsennrich/wmt16_neural/en-de/exp1/model.iter{54,57,60,63}0000.npz
31.5 & 37.5\\ % /mnt/hoenir0/rsennrich/wmt16_neural/de-en/exp1/model.iter{60,63,66,69}0000.npz
+r2l reranking & \textbf{28.1} & \textbf{34.2} &%/mnt/hoenir0/rsennrich/wmt16_neural/en-de/exp1/model.iter{54,57,60,63}0000.npz, /mnt/hoenir0/rsennrich/wmt16_neural/en-de/exp6/model.iter{60,63,66,69}0000.npz
\textbf{32.1} & \textbf{38.6}\\ %/mnt/hoenir0/rsennrich/wmt16_neural/de-en/exp1/model.iter{60,63,66,69}0000.npz, /mnt/hoenir0/rsennrich/wmt16_neural/de-en/exp2/model.iter{51,54,57,60}0000.npz
\end{tabular}
\caption{English$\leftrightarrow$German translation results ({\sc Bleu}) on dev (newstest2015) and test (newstest2016). Submitted system in bold.}
\label{results-de}
\end{table}

Table \ref{results-de} shows results for English$\leftrightarrow$German.
We observe improvements of 3.4--5.7 {\sc Bleu} from training with a mix of parallel and synthetic data, compared to the baseline that is only trained on parallel data.
Using an ensemble of the last 4 checkpoints gives further improvements (1.3--1.7 {\sc Bleu}).
Our submitted system includes reranking of the 50-best output of the left-to-right model with a right-to-left model -- again an ensemble of the last 4 checkpoints -- with uniform weights.
This yields an improvements of 0.6--1.1 {\sc Bleu}.

\subsection{English$\leftrightarrow$Czech}
For English$\rightarrow$Czech, we trained our baseline model on the complete WMT16 parallel training set (including CzEng 1.6pre \cite{czeng16:2016}), until we observed convergence on our
heldout set (newstest2014). This took approximately 1M minibatches, or 3 weeks. Then we continued training the model on a new parallel corpus, comprising
8.2M sentences back-translated from the Czech monolingual news2015, 5 copies of news-commentary v11, and 9M sentences sampled from Czeng 1.6pre. The
model used for back-translation was a neural MT model from earlier experiments, trained on WMT15 data. The training
on this synthetic mix continued for a further 400,000 minibatches.

The right-left model was trained using a similar process, but with the target side of the parallel corpus reversed prior to training. The
resulting model had a slightly lower {\sc Bleu} score on the dev data than the standard left-right model.
We can see in Table \ref{results-cs} that back-translation improves performance by 
2.2--2.8 {\sc Bleu}, and that 
 the final system (+r2l reranking) improves by 0.7--1.0 {\sc Bleu} on the ensemble of 4, and 4.3--4.9 on the baseline.

For Czech$\rightarrow$English the training process was similar to the above, except that we created the synthetic training data (back-translated from samples of news2015 monolingual English)
in batches of 2.5M, and so were able to observe the effect of increasing the amount of synthetic data. After training a baseline model on all the WMT16 parallel set, 
we continued training with a parallel corpus consisting of 2 copies of the 2.5M sentences of back-translated data, 5 copies of news-commentary v11, and a matching 
quantity of data sampled from Czeng 1.6pre. After training this to convergence, we restarted training from the baseline model using 5M sentences of 
back-translated data, 5 copies of news-commentary v11, and a matching quantity of data sampled from Czeng 1.6pre. We repeated this with 7.5M sentences from
news2015 monolingual, and then with 10M sentences of news2015. The back-translations were, as for English$\rightarrow$Czech, created with an earlier NMT 
model trained on WMT15 data.
Our final Czech$\rightarrow$English was an ensemble of 8 systems -- the last 4 save-points of the 10M synthetic data run, and the last 4 save-points of the
7.5M run. We show this as ensemble8 in Table \ref{results-cs}, and the +synthetic results are on the last (i.e. 10M) synthetic data run. 

We also show in Table \ref{increase-bt} how increasing the amount of back-translated data
affects the results. We see that most of the gain from back-translation comes with the first batch, but increasing the amount of back-translated data does
gradually improve performance.

\begin{table}
\centering
\begin{tabular}{l|cc|cc}
system & \multicolumn{2}{c|}{EN$\to$CS} & \multicolumn{2}{c}{CS$\to$EN}\\
& dev & test & dev & test\\
\hline
baseline & 18.5 & 20.9 &   %/mnt/meili0/bhaddow/experiments/wmt16/en-cs/run01/model.npz.npz.best_bleu
23.8 & 25.3\\ %/mnt/meili0/bhaddow/experiments/wmt16/cs-en/run01/model.npz.npz.best_bleu
+synthetic & 20.7 & 23.7 &  %/mnt/meili0/bhaddow/experiments/wmt16/en-cs/run02/model.npz.npz.best_bleu
27.2 & 30.1 \\ %/mnt/meili0/bhaddow/experiments/wmt16/cs-en/run05/model.npz.npz.best_bleu
+ensemble & 22.1 & 24.8 &  %/mnt/meili0/bhaddow/experiments/wmt16/en-cs/run02/model.iter{147,144,141,138}0000.npz
28.6 & 31.0 \\ %/mnt/meili0/bhaddow/experiments/wmt16/cs-en/run05/model.iter{141,144,147,150}0000.npz
+ensemble8 & -- & -- & 
\textbf{29.0} & \textbf{31.4} \\ %/mnt/meili0/bhaddow/experiments/wmt16/cs-en/run05/model.iter{141,144,147,150}0000.npz, %/mnt/meili0/bhaddow/experiments/wmt16/cs-en/run04/model.iter{147,150,153,156}0000.npz
+r2l reranking & \textbf{22.8} & \textbf{25.8} &  %/mnt/meili0/bhaddow/experiments/wmt16/en-cs/run02/model.iter{147,144,141,138}0000.npz, /mnt/meili0/bhaddow/experiments/wmt16/en-cs/run04/model.iter{153,156,159,162}0000.npz
-- & -- \\
\end{tabular}
\caption{English$\leftrightarrow$Czech translation results ({\sc Bleu}) on dev (newstest2015) and test (newstest2016). Submitted system in bold.}
\label{results-cs}
\end{table}

\begin{table}
\centering
\begin{tabular}{l|cc|cc}
system & \multicolumn{2}{c|}{best single} & \multicolumn{2}{c}{ensemble4}\\
& dev & test & dev & test\\
\hline
baseline & 23.8 & 25.3 &   %/mnt/meili0/bhaddow/experiments/wmt16/cs-en/run01/model.npz.npz.best_bleu
25.5 & 26.8 \\  %/mnt/meili0/bhaddow/experiments/wmt16/cs-en/run01/model.iter{105,102,99,86}.npz
+2.5M synthetic & 26.7 & 29.4 &  %/mnt/meili0/bhaddow/experiments/wmt16/cs-en/run02/model.npz.npz.best_bleu
27.7 & 30.4 \\ %/mnt/meili0/bhaddow/experiments/wmt16/cs-en/run02/model.iter{120,117,114,111}.npz
+5M synthetic & 27.2 & 29.3 &  %/mnt/meili0/bhaddow/experiments/wmt16/cs-en/run03/model.npz.npz.best_bleu
28.2 & 30.4 \\ %/mnt/meili0/bhaddow/experiments/wmt16/cs-en/run03/model.iter{144,141,138,135}.npz
+7.5M synthetic & 27.2 & 29.7 &  %/mnt/meili0/bhaddow/experiments/wmt16/cs-en/run04/model.npz.npz.best_bleu
28.4  & 30.8 \\ %/mnt/meili0/bhaddow/experiments/wmt16/cs-en/run04/model.iter{156,153,150,147}.npz
+10M synthetic & 27.2 & 30.1 &  %/mnt/meili0/bhaddow/experiments/wmt16/cs-en/run05/model.npz.npz.best_bleu
28.6 & 31.0 \\ %/mnt/meili0/bhaddow/experiments/wmt16/cs-en/run05/model.iter{150,147,144,141}.npz
\end{tabular}
\caption{Czech$\rightarrow$English translation results ({\sc Bleu}) on dev (newstest2015) and test (newstest2016), after continued training with increasing amounts of 
back-translated synthetic data. For each row, training was continued from the baseline model until convergence.}
\label{increase-bt}
\end{table}

\subsection{English$\leftrightarrow$Romanian}

\begin{table}[t]
\centering
\begin{tabular}{l|cc|cc}
system & \multicolumn{2}{c|}{EN$\to$RO} & \multicolumn{2}{c}{RO$\to$EN}\\
& dev & test & dev & test\\
\hline
baseline & 20.2 & 19.2 &%/mnt/hoenir0/rsennrich/wmt16_neural/en-ro/exp1/model.npz.npz_best_bleu
23.6 & 22.7\\ %/mnt/hoenir0/rsennrich/wmt16_neural/ro-en/lexi/run01/model.npz.npz_best_bleu
+dropout & 24.2 & 23.9 & %/mnt/hoenir0/rsennrich/wmt16_neural/en-ro/exp3/model.npz.npz.best_bleu
28.7 & 27.8\\ %/mnt/hoenir0/rsennrich/wmt16_neural/ro-en/exp1/model.npz.npz_best_bleu
+remove diacritics & - & - & 30.0 & 29.2\\ %/mnt/hoenir0/rsennrich/wmt16_neural/ro-en/exp2/model.npz.npz.best_bleu
+synthetic & \textbf{29.3} & \textbf{28.1} & %/mnt/hoenir0/rsennrich/wmt16_neural/en-ro/exp4/model.npz.npz.best_bleu
34.8 & 33.3\\ %/mnt/hoenir0/rsennrich/wmt16_neural/ro-en/exp5/model.npz.npz.best_bleu
+ensemble & 29.0 & 28.2 & %/mnt/hoenir0/rsennrich/wmt16_neural/en-ro/exp4/model.iter{63,66,69,72}0000.npz
\textbf{35.3} & \textbf{33.9}\\ %/mnt/hoenir0/rsennrich/wmt16_neural/ro-en/exp5/model.iter{48,51,54,57}0000.npz
\end{tabular}
\caption{English$\leftrightarrow$Romanian translation results ({\sc Bleu}) on dev (newsdev2016), and test (newstest2016). Submitted system in bold.}
\label{results-ro}
\end{table}

The results of our English$\leftrightarrow$Romanian experiments are shown in Table \ref{results-ro}.
This language pair has the smallest amount of parallel training data, and we found dropout to be very effective, yielding improvements of 4--5 {\sc Bleu}.\footnote{We also tested dropout for EN$\to$DE with 8 million sentence pairs of training data, but found no improvement after 10 days of training. We speculate that dropout could still be helpful for datasets of this size with longer training times and/or larger networks.}

We found that the use of diacritics was inconsistent in the Romanian training (and development) data,  so
for Romanian$\to$English we removed diacritics from the Romanian source side, obtaining improvements of 1.3--1.4 {\sc Bleu}.

Synthetic training data gives improvements of 4.1--5.1 {\sc Bleu}.
for English$\to$Romanian, we found that the best single system outperformed the ensemble of the last 4 checkpoints on dev, and we thus submitted the best single system as primary system.

\subsection{English$\leftrightarrow$Russian}

\begin{table}
\centering
\begin{tabular}{l|cc|cc}
system & \multicolumn{2}{c|}{EN$\to$RU} & \multicolumn{2}{c}{RU$\to$EN}\\
& dev & test & dev & test\\
\hline
baseline & 21.3 & 20.3 &%/mnt/hoenir0/rsennrich/wmt16_neural/en-ru/exp1/model.npz.npz_best_bleu
22.7 & 22.5 \\ %/mnt/hoenir0/rsennrich/wmt16_neural/ru-en/exp1/model.npz.npz_best_bleu
+synthetic & 25.8& 24.3 &% %/mnt/hoenir0/rsennrich/wmt16_neural/en-ru/exp2/model.npz.npz_best_bleu
27.1 & 26.9\\ %/mnt/hoenir0/rsennrich/wmt16_neural/ru-en/exp2/model.npz.npz_best_bleu
+ensemble & \textbf{27.0} & \textbf{26.0} &%/mnt/hoenir0/rsennrich/wmt16_neural/en-ru/exp2/model.iter{51,54,57,60}0000.npz
\textbf{28.3} & \textbf{28.0}\\ %/mnt/hoenir0/rsennrich/wmt16_neural/ru-en/exp2/model.iter{51,54,57,60}0000.npz
\end{tabular}
\caption{English$\leftrightarrow$Russian translation results ({\sc Bleu}) on dev (newstest2015) and test (newstest2016). Submitted system in bold.}
\label{results-ru}
\end{table}

For English$\leftrightarrow$Russian, we cannot effectively learn BPE on the joint vocabulary because alphabets differ.
We thus follow the approach described in \cite{DBLP:journals/corr/SennrichHB15}, first mapping the Russian text into Latin characters via ISO-9 transliteration,
then learning the BPE operations on the concatenation of the English and latinized Russian training data, then mapping the BPE operations back into Cyrillic alphabet.
We apply the Latin BPE operations to the English data (training data and input), and both the Cyrillic and Latin BPE operations to the Russian data.

Translation results are shown in Table \ref{results-ru}.
As for the other language pairs, we observe strong improvements from synthetic training data (4--4.4 {\sc Bleu}).
Ensembles yield another 1.1--1.7 {\sc Bleu}.

\section{Shared Task Results}

\begin{table}
\centering
\begin{tabular}{c|c|cc}
direction & {\sc Bleu} rank & \multicolumn{2}{c}{human rank}\\
\hline
EN$\to$CS & 1 of \phantom{0}9 & 1 &of 20\\
EN$\to$DE & 1 of 11 & 1 &of 15\\
EN$\to$RO & 2 of 10 & 1--2 &of 12 \\
EN$\to$RU & 1 of \phantom{0}8 & 2--5 &of 12\\
\hline
CS$\to$EN & 1 of \phantom{0}4 & 1 &of 12\\
DE$\to$EN & 1 of \phantom{0}6 & 1 &of 10\\
RO$\to$EN & 2 of \phantom{0}5 & 2 &of \phantom{0}7\\
RU$\to$EN & 3 of \phantom{0}6 & 5 &of 10\\
\end{tabular}
\caption{Automatic ({\sc Bleu}) and human ranking of our submitted systems (uedin-nmt) at WMT16 shared news translation task. Automatic rankings are taken from \url{http://matrix.statmt.org} , only considering primary systems. Human rankings include anonymous online systems, and for EN$\leftrightarrow$CS, systems from the tuning task.}
\label{ranking}
\end{table}

Table \ref{ranking} shows the ranking of our submitted systems at the WMT16 shared news translation task.
Our submissions are ranked (tied) first for 5 out of 8 translation directions in which we participated: EN$\leftrightarrow$CS, EN$\leftrightarrow$DE, and EN$\to$RO.
They are also the (tied) best constrained system for EN$\to$RU and RO$\to$EN, or 7 out of 8 translation directions in total.

Our models are also used in QT21-HimL-SysComb \cite{qt21syscomb2016}, ranked 1--2 for EN$\to$RO, and in AMU-UEDIN \cite{junczys2016}, ranked 2--3 for EN$\to$RU, and 1--2 for RU$\to$EN.

\section{Conclusion}

We describe Edinburgh's neural machine translation systems for the WMT16 shared news translation task.
For all translation directions, we observe large improvements in translation quality from using synthetic parallel training data, obtained by back-translating in-domain monolingual target-side data.
Pervasive dropout on all layers was used for English$\leftrightarrow$Romanian, and gave substantial improvements.
For English$\leftrightarrow$German and English$\to$Czech, we trained a right-to-left model with reversed target side, and we found reranking the system output with these reversed models helpful.

\section*{Acknowledgments}

This project has received funding from the European Union's Horizon 2020 research and innovation
programme under grant agreements 645452 (QT21), 644333 (TraMOOC) and 644402 (HimL).

\bibliographystyle{acl2016}
\bibliography{wmt16}

\end{document}